\documentclass[journal,comsoc]{IEEEtran}
\usepackage{amsmath,amsfonts}
\usepackage{algorithmic}
\usepackage{algorithm}
\usepackage{array}
\usepackage{color}
\usepackage[caption=false,font=normalsize,labelfont=sf,textfont=sf]{subfig}
\usepackage{textcomp}
\usepackage{stfloats}
\usepackage{url}
\usepackage{verbatim}
\usepackage{graphicx}
\usepackage{cite}
\usepackage[utf8]{inputenc}
\usepackage[english]{babel}
\usepackage{multirow}
\usepackage[backref]{hyperref}

\begin{document}

\title{A Specific Task-oriented Semantic Image Communication System for substation patrol inspection}

\author{Senran Fan, Haotai Liang, Chen Dong*, Xiaodong Xu, Geng Liu
\thanks{Senran Fan, Haotai Liang are with the State Key Laboratory of Networking and Switching Technology, Beijing University of Posts and Telecommunications, Beijing, 100876, China. (E-mail: FSR@bupt.edu.cn; lianghaotai@bupt.edu.cn)}
\thanks{Xiaodong Xu is with the State Key Laboratory of Networking and Switching Technology, Beijing University of Posts and Telecommunications, Beijing, China, and also with the Department of Broad-band Communication, Peng Cheng Laboratory, Shenzhen, Guangdong, China. (E-mail: xuxiaodong@bupt.edu.cn)}
\thanks{Geng Liu is with the Beijing Smart-chip Microelectronics Technology Co.,Ltd. (E-mail: liugeng@sgchip.sgcc.com.cn)}
\thanks{*Chen Dong is the corresponding author and with the State Key Laboratory of Networking and Switching Technology, Beijing University of Posts and Telecommunications, Beijing, 100876, China. (E-mail: dongchen@bupt.edu.cn)}}

\markboth{}
{}

\maketitle

\begin{abstract}
    Intelligent inspection robots are widely used in substation patrol inspections to identify potential safety hazards by patrolling substations and sending back scene images. However, in areas with weak signals, the scene images may not be successfully transmitted, thereby reducing the quality of the robots’ work. 
    To reliably transmit high quality inspection images in weak signal areas, a Specific Task-oriented Semantic Communication System for Images (STSCI) is proposed. This system involves the extraction, transmission, restoration, and enhancement of semantic features to obtain clearer images sent by intelligent robots under weak signals. Inspired by the fact that only specific details of the image are required in substation patrol inspection tasks, we propose a new paradigm of semantic enhancement to ensure the clarity of key semantic information under low signal-to-noise ratio conditions. Experiments demonstrate that our STSCI generally outperforms traditional image-compression-based and channel-coding-based as well as other semantic communication systems in substation patrol inspection tasks under low signal-to-noise ratio conditions.
\end{abstract}


\section{Introduction}
\IEEEPARstart{S}ubstation inspections are critical in ensuring the safe and reliable operation of the power grid. However, weak signal areas within substations pose a significant challenge for inspection, as they can severely affect the quality of the inspection images. This is a common issue faced by most power utilities and can be caused by various factors such as signal obstruction from metal equipment, signal attenuation due to channel fading, weather conditions, and equipment aging. Thus, addressing this issue is of paramount importance, and many studies have attempted to tackle the relevant problem, such as\cite{SCN, Wireless}. However, these studies were all based on traditional communication methods. The traditional communication system is reaching Shannon’s Limit in physical layers, and the channel coding method relied upon by traditional communication systems performs poorly under low signal-to-noise conditions, leading to unsatisfactory transmission performance, especially in weak signal areas.\par
To overcome this challenge, we propose a Specific Task-oriented Semantic Communication System for Images (STSCI), which fully utilizes the characteristics of specific inspection tasks, such as fixed image sources, fixed channel conditions and focusing on the key semantic content in the images. The STSCI combines semantic communication, deep learning, and Joint Source-Channel Coding(JSCC)\cite{deepJSCC} to address the problem of normal inspection in weak signal areas of substations.\par  
Semantic communication systems are considered a new and promising direction in the communication fields. First mentioned in Shannon and Weaver's paper\cite{shannon}, semantic-based communication systems achieve extreme data compression by only focusing on key semantic information. In specific communication scenarios, a large amount of task-irrelevant information is involved in transmitting information, leading to a waste of communication resources. Particularly for the task of intelligent substation patrol inspection, what really matters are only the key semantic contents, such as the areas with key units of the image. As introduced in\cite{SC,6G}, the semantic communication system achieves highly efficient communication by using semantic feature extraction, transmission, and reconstruction to compress the image while retaining the key semantic information.\par
Deep learning can extract the semantic features from the image. Semantic segmentation networks\cite{FCN,Segnet,Unet} as well as target detection networks\cite{FRCNN,SSD,yolov1,yolov2} are powerful in semantic features extracting and analysis. Additionally, thanks to the sophisticated design of adversarial loss functions, GAN networks\cite{GAN} are capable of generating images based on semantic features. In unsupervised fields, the auto-encoder\cite{Autoencoder} is an inspiring architecture for semantic feature extraction as it compresses high-dimensional data into a low-dimensional encoding that is used for data reconstruction. By combing GAN networks and the structure of auto-encoder, the system can decrease distortion in semantic contents of images during the process of extreme compression as well as transmission. \par
The traditional communication systems rely on source coding and channel coding. We propose replacing the source coding with GAN networks and using Joint Source-Channel Coding (JSCC)\cite{deepJSCC,deepJSCC-f} to resist noise in channels. By leading the simulated-channel models into the training process of deep networks, the networks can perform better in real-world channel conditions especially under low signal-to-noise ratio situations. \par
In fact, a layer-based semantic communication system for images (LSCI) proposed in\cite{LSCI} combines semantic communication techniques with GAN network to encode and decode information, demonstrating excellent performance in image transmission. This served as a great inspiration for our research.\par
Building upon the above studies, the STSCI has more powerful network architectures and fully utilizes the characteristics of specific inspection tasks, which can achieve highly effient communication while minimizing the loss of key semantic information even under low signal-to-noise conditions. The system is a GAN-based auto-encoder-structured network using RRDB blocks as well as tricks from conditional GAN\cite{conditional} for image compression, transmission, and reconstruction. In addition, a YOLONet is involved to locate the image's specific semantic contents, which will then be embedded and sent to the enhancement CNN model to improve the transmission quality of the important semantic contents of the images, to make sure there are no errors or missing when making security checks with the transmitted images. The key contributions of this paper are summarized as follows.
\begin{figure*}[ht]
    \centering
    \includegraphics[scale=0.31]{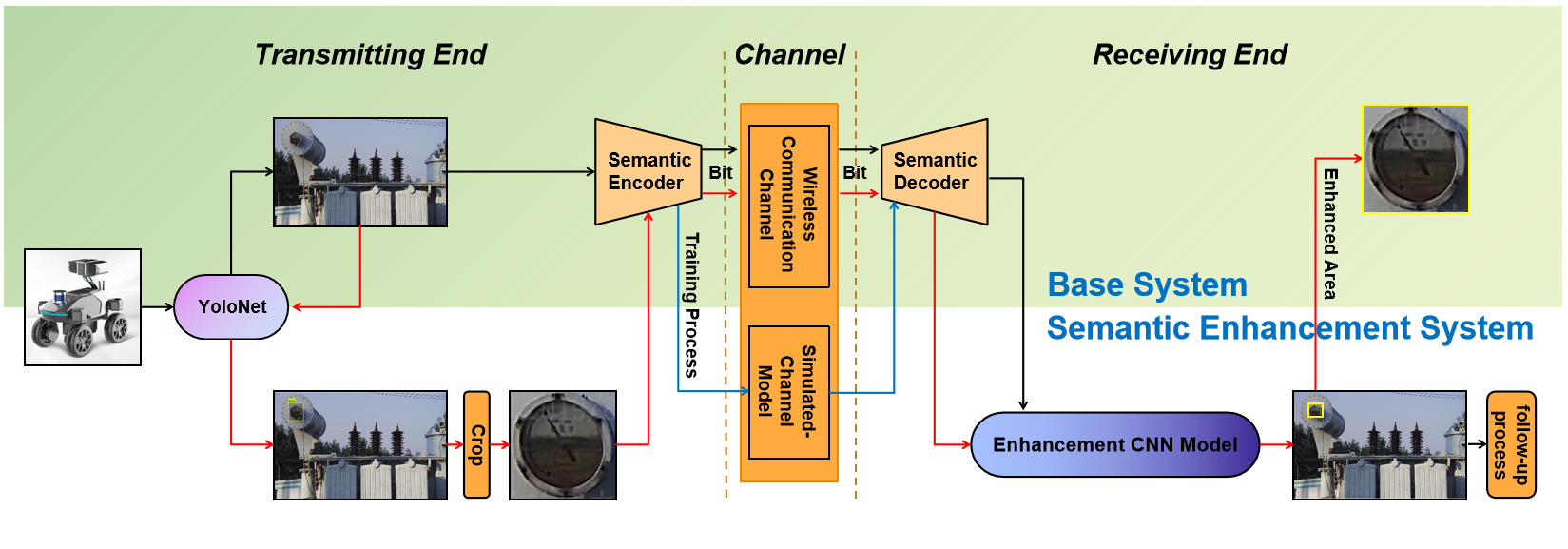}
    \caption{The framework of STSCI. STSCI consists of two systems, the base system and the semantic enhancement system. The base system consists of a semantic encoder, a semantic decoder, and a simulated-channel model (trained only), with the process indicated by black lines. The semantic enhancement system with the process indicated by red lines, on the other hand, includes a YOLONet for identifying key semantic content and an enhancement CNN network that utilizes extra information to enhance the transmission quality of the key semantic information. The simulated-channel model is only used during the model training process to simulate a real-world wireless channel. This process is indicated by blue lines.}
\end{figure*}  
\begin{itemize}
    \item[(1)]
A specific task-oriented semantic communication system for image is proposed for the transmission of images obtained by intelligent robots in the substation patrol inspection task. A new paradigm of key semantic
contents extraction and preservation for such specific tasks is proposed. A YOLONet is involved to locate the key semantic contents, which are then sent to an enhancement CNN model to improve the transmission quality of those areas. 
\end{itemize}
\begin{itemize}
    \item[(2)]
A GAN-based auto-encoder structure network is designed. Combined with RRDB blocks, ChannelNorm, idea of conditional GAN and some other tricks, the network can extremely compress the images into the semantic feature encoding and reconstruct them after the transmission.                                     
\end{itemize}
\begin{itemize}
    \item[(3)]
Through simulations and experiments, this paper demonstrates the efficacy of the semantic communication system in handling the specific tasks. The results demonstrate that the STSCI outperforms traditional communication systems in tasks with fixed image source and fixed channel conditions. Specifically, the STSCI provides better transmission quality, especially under low signal-to-noise ratio channel conditions. This significantly expands the effective signal coverage area, which ensures the proper functioning of the intelligent robots when patrolling the weak signal areas of the substation.
\end{itemize}
This paper is organized as follows. Section II provides an overview of the the Specific Task-oriented Semantic Communication System for Images (STSCI), including the system workflow, model architectures and training flow path for the two parts of the STSCI. In Section III, we present a direct comparison between the STSCI and other image communication systems to quantify the performance of the STSCI using the proposed method. Finally, we draw conclusions in Section IV.

\begin{figure*}[ht]
    \centering
    \includegraphics[scale=0.35]{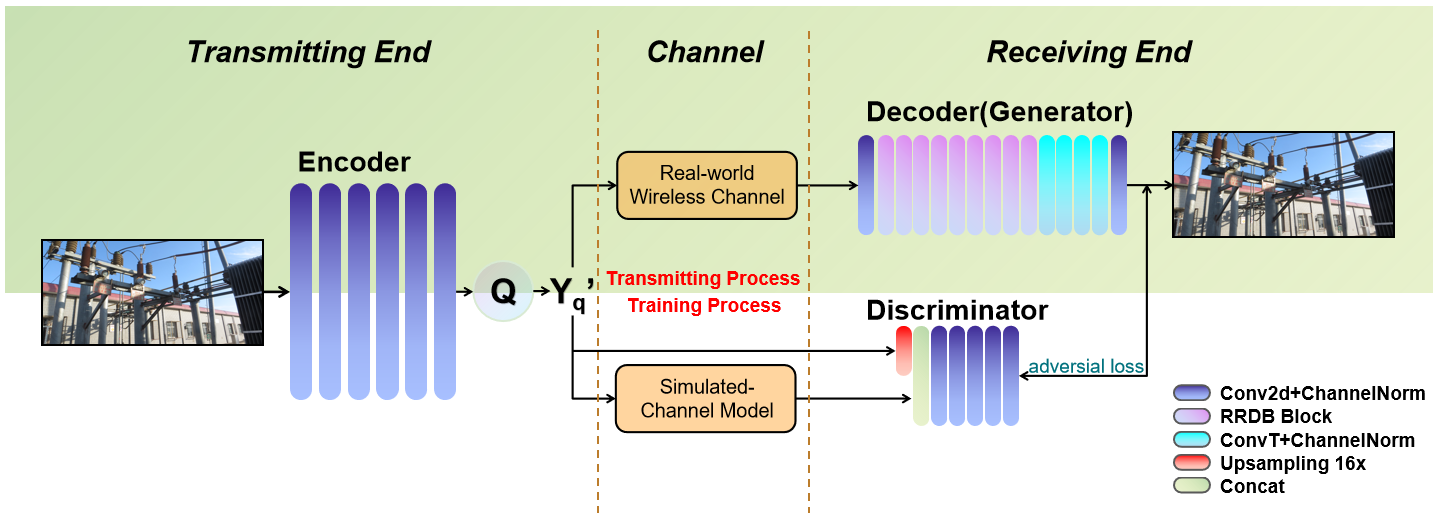}
    \caption{The architecture of the base system, including the encoder at transmitting end and the decoder at receiving end. The modules with green background describes the process of image transmission. During training, a simulated-channel model is added for source-channel joint encoding, as well as a discriminator in the GAN model to assist in training the generator, which is the decoder. The networks in the diagram are represented by several rectangular blocks, with the number of blocks matching the number of network layers. Blocks of different colors represent different network layers.}
\end{figure*}  

\section{Specifc task-oriented semantic communication system}

As shown in Fig. 1, the Specific Task-Oriented Semantic Communication System for Images (STSCI) consists of two parallel parts: the base system and the semantic enhancement system. \par 
The base system is a GAN-based auto-encoder network, containing a semantic encoder at the transmitting end and a semantic decoder at the receiving end. The encoder compresses the captured inspection images into a semantic encoding containing semantic features, which is transmitted over the channel to the receiving end and then reconstructed into a image by the decoder. \par
Meanwhile, the semantic enhancement system consists of a YOLONet at the transmitting end to identify areas with key semantic contents in the image and an enhancement CNN model at the receiving end that generates high-quality images and improves the clarity and quality of areas with key semantic contents. \par
The two parts together form the communication system for encoding, transmitting and decoding the inspection images. The overall workflow of the system is shown in Fig. 1, the inspection images captured by the robot are first fully processed by the basic system, encoded into semantic encoding by the encoder,  transmitted to the receiving end via the wireless channel, and reconstructed into an image by the GAN-based decoder at the receiving end. At the same time, at the transmitting end, the YOLONet is used to identify and locate key semantic content areas in the image, the located areas will be encoded and transmitted separately through the basic system. Therefore, the transmitting end will obtain not only the complete inspection image, but also the sub-image corresponding to the key semantic content areas. Then the enhancement CNN model takes these two images as input and outputs a complete image with the key semantic parts more precise and clear. \par
In the following text, both parts of the STSCI will be introduced in detail. \par
~\\
\emph{A. Base System}\par
As shown in Fig. 2, the base system is a deep neural network consisting of an encoder network, simulated-channel models, and a GAN-based decoder network. At the transmitting end, the encoder compresses the obtained inspection images into semantic encoding that are transmitted through the channel and reconstructed by the decoder at the receiving end. The simulated-channel models and the discriminator in GAN-based decoder network are only used during the training process. \par 
The key design of the base system is using a GAN network to reconstruct the image from the semantic encoding instead of a traditional CNN model in a general auto-encoder structure. GAN networks utilize adversarial training to guide the generator to effectively learn the image distribution and approach the real images. Their wide application in image enhancement tasks such as image denoising and super-resolution\cite{srgan, esrgan} has proven their extraordinary performance in high-quality reconstruction. \par
The generator backbone uses Residual-in-Residual Dense Blocks (RRDBs), inspired by ESRGAN\cite{esrgan}, a seminal work in single image super resolution with excellent generated results. RRDB combines the advantages of residual learning and dense connections. Each RRDB contains multiple densely connected residual blocks, which enables features to be fully reused and propagated efficiently within each RRDB to extract richer features and utilize correlations between channels. Compared to standard residual blocks, RRDB achieve superior performance with significantly fewer parameters. \par
The system employs ChannelNorm in the generator, first mentioned in\cite{hific}, it performs normalization across the channel dimension instead of the batch or spatial dimensions like Batch Norm and Instance Norm. It encourages competition between channels, leading to more discriminative feature representations. The richer channel-wise feature representations significantly benefit high-resolution image generation tasks. \par
Inspired by\cite{hific}, the discriminator shares the structure of that in conditional GAN. The discriminator receives not only the generated images and real images but also the corresponding semantic encoding, forcing attention the connections between the semantic encoding and the image as well as the difference between images with different semantic encoding. This helps the adversarial loss covers more useful information to improve reconstruction quality. \par
Considering that the structure of the auto-encoder involved in the semantic communication system is highly consistent with the information communication process, Joint Source-Channel Coding (JSCC) was proposed in\cite{deepJSCC}. Instead of using additional channel coding like LDPC, JSCC resists noise in channels by adding noise through simulated-channel models when training auto-encoder networks. This forms an anti-noise communication system that ensures high-quality image transmission under low signal-to-noise ratio conditions. JSCC methods have limitations in being constrained by fixed data sources and channel conditions, which lead to the lack of generalization in deep-based semantic communication systems. However, such constraints can be ignored in this task since the information source and channels are fixed.\par
Regarding the loss functions that play a decisive role in training the networks, Mean Squared Error (MSE) loss and Learned Perceptual Image Patch Similarity (LPIPS) loss are chosen to measure the distortion between the original images and the generated ones. MSE loss measures the difference per pixel and shows their distance in the high-dimensional space, which helps maintain the similarity. Meanwhile, LPIPS loss proposed in\cite{perceptual} is calculated through a VGG-net that has been trained previously. The pre-trained VGG-net gives more attention to the structure and texture of the images and does well in telling such differences between images. LPIPS loss helps fill this gap and makes the generated images closer to the original ones in visual. \par 
According to all the above introductions and the structure shown in Fig. 2, the complete process of the base system is as follows. \par
The image \mbox{$X$} to be transmitted is sent in to the encoder first to get the semantic features \mbox{$Y$},\par
\begin{equation}\label{eqn-1} 
    Y = E(x).
\end{equation}
The nearest neighbor quantization operation is then performed on the extracted semantic features \mbox{$Y$}, \par 
\begin{equation}\label{eqn-1} 
    Y_q(i) = argmin_j||Y_i-I_j||. 
\end{equation}
Where the set \mbox{$I$} of quantization centers is: \par
\begin{equation}\label{eqn-1} 
    I = \{I_0, I_1, ..., I_j , ..., I_l\}. 
\end{equation}
According to JSCC, then the quantized semantic feature \mbox{$Y_q$} is sent to the simulated-channel models. In this paper, AWGN model is chosen as the simulated-channel model,\par
\begin{equation}\label{eqn-1} 
    {Y_q}^{'} = h \cdot Y_q + n. 
\end{equation}
In this formula, \mbox{$h$} represents the channel gain, while \mbox{$n$} represents the independent identically distributed Gaussian noise. Such model simulates the feature's distortion transmitted in the real-world channel and gives the base system the ability to resist the noise.\par
The image \mbox{$X^{'}$} is generated by the generator(the decoder network) from the processed latent \mbox{${Y_q}^{'}$} at the receiving end,\par
\begin{equation}\label{eqn-1} 
    X^{'} = G({Y_q}^{'}).
\end{equation}
The encoder maps the source image \mbox{$X$} to a specific distribution \mbox{$P_X$}. The generator G tries to map samples \mbox{$Y$} from a fixed known distribution \mbox{$P_Y$} to \mbox{$P_X$}, while the discriminator D is learned to tell the difference between such two distributions using the sampled data \mbox{$X$} and the generated \mbox{$X^{'}$}. A properly trained discriminator helps the generator to find and simulate the distribution \mbox{$P_X$} more preciously. Involving the idea of conditional GANs as mentioned before, the adversarial loss is as follows.\par
\begin{equation}\label{eqn-1} 
    L_G = -log(D(X^{'}, {Y_q}^{'})),
\end{equation}
\begin{equation}\label{eqn-1} 
    L_D = -log(1-D(X^{'}, {Y_q}^{'})) - log(D(X, {Y_q}^{'})).
\end{equation}
Besides, when optimizing the encoder and the generator, the MSE loss and the LPIPS loss are also involved to measure the texture and perception distance between the source image X and the generator image \mbox{$X^{'}$}. Moverover, helping to initialize these two networks, these two kinds of loss guide the generator and discriminator to be trained on the right direction. So the final loss for the encoder and the generator are as follows.\par
In the initial training:\par
\begin{equation}\label{eqn-1} 
    L_{EG} = ||X-X^{'}|| + \alpha LPIPS(X, X^{'}).
\end{equation} \par
In the final training: \par
\begin{equation}\label{eqn-1} 
    L_{EG} = ||X-X^{'}|| + \alpha LPIPS(X, X^{'}) +  \beta [-log(D(X^{'}, {Y_q}^{'})]. 
\end{equation}
Where \mbox{$\alpha$} and \mbox{$\beta$} are coefficient weights used when calculating the loss function, to adjust the influence of different loss functions on the total loss function. \par
\begin{figure*}[ht]
    \centering
    \includegraphics[scale=0.30]{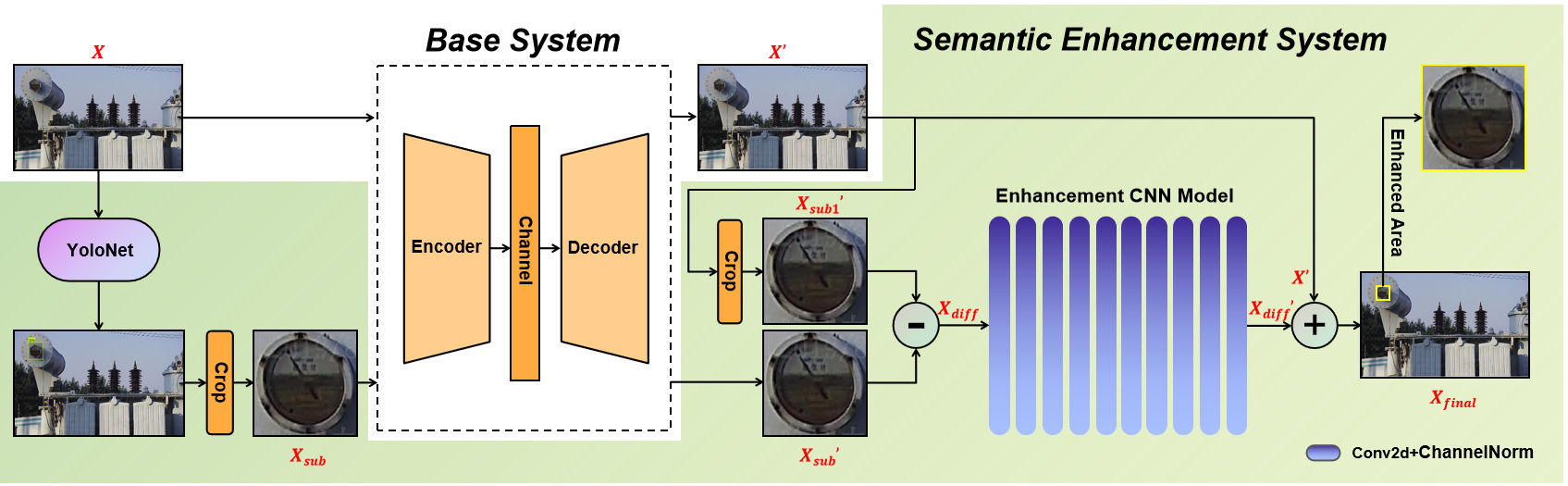}
    \caption{The process of the semantic enhancement system. The YOLONet is used to locate areas in the image that contain key semantic information. Those areas are then cropped into sub-images and transmitted to the receiving end through the basic system. This portion, serving as additional information, is combined with the complete image transmitted from the basic system and input into the enhancement CNN model to obtain the final image.}
\end{figure*} 
~\\

\noindent \emph{B. Semantic Enhancement System}\par

The semantic ehancement system is designed to improve the transmission quality of the key semantic content that is important for specific tasks, such as the panels or electrical insulators in intelligent substation patrol inspections. The system consists of two parts: a YOLONet that locates the area with key semantic content, which will be sent to the base system, and an enhancement CNN model that generates high-quality images at the receiving end using the transmitted image and the areas with key semantic content as input.   \par
Target detection network YOLONet is involved in our system to locate the key semantic contents instead of semantic segmentation networks such as U-net or FCN, as used in some other semantic communication systems like\cite{LSCI}. For the reason that 
the pre-trained YOLONet has the ability to find and locate objects that need to be captured during the patrol task, so this network can not only locate and mark the area containing key semantic information during the semantic communication process, but can also help the intelligent robots to determine whether there are objects in the patrol list that need to be captured, and how to adjust the position, angle, and focal length of the camera to obtain a sharper image. Considering the constraint of storage space in the patrol robot, using YOLONet, which can perform multiple tasks, is a cost-effective choice.\par
In this paper, YOLOv5s is used, which has the advantages of being faster, smaller, and more accurate compared to the previous versions of YOLO.  The "s" in YOLOv5s stands for "small", indicating that this variant has a smaller model size and is optimized for running on devices with limited computational resources. Despite its smaller size, YOLOv5s still achieves high accuracy in object detection tasks. \par

As shown in Fig. 3, semantic enhancement system's process is as follows.\par
The area \mbox{$X_{sub}$} with key semantic contents is located by the YOLONet with input of source image \mbox{$X$},\par
\begin{equation}\label{eqn-1} 
    X_{sub} = YOLONet(X).
\end{equation}
After sent into the base system, \mbox{$X_{sub}$} is encoded, transmitted and finally reconstructed as the \mbox{${X_{sub}}^{'}$} at the receiving end,\par
\begin{equation}\label{eqn-1} 
    {X_{sub}}^{'} = Base\_system(X_{sub}).
\end{equation}
At the same time, the whole image \mbox{$X$} is transmitted through the base system to get another area \mbox{${X_{sub1}}^{'}$} with key semantic contents cut from the reconstructed image \mbox{$X^{'}$}. The difference between these two sub-images is calculated as follows.\par
\begin{equation}\label{eqn-1} 
    X_{diff} = {X_{sub1}}^{'} - {X_{sub}}^{'}.
\end{equation}
The DIFF image is sent to the enhancement CNN model whose job is to balance the difference between two sub-image to make full use of these extra information to let the transmitted image as close as the original one in the area with key semantic contents, 
\begin{equation}\label{eqn-1} 
    {X_{diff}}^{'} = Enhancement\_CNN\_Model(X_{diff}).
\end{equation} 	
The final image is formed as follows.\par
\begin{equation}\label{eqn-1} 
    X_{final} =  X^{'} + {X_{diff}}^{'}.
\end{equation} 	
In this task, the similarity between the final image \mbox{$X_{final}$} and the original image \mbox{$X$} is focused on, which can help decrease the possibility of errors or missing during analyzing the images. So we choose the MSE loss and SSIM loss to optimize the enhancement CNN models, and parameters in the YOLONet as well as the base model are fixed during the optimization,
\begin{equation}\label{eqn-1} 
    L_{enhancement} = ||X - X_{final}|| + \alpha SSIM(X, X_{final}).
\end{equation}
Where \mbox{$\alpha$} is coefficient weights used when calculating the loss function, to adjust the influence of different loss functions on the total loss function. \par
The details of networks involved in the STSCI is shown in Table I and Table II.\par
\begin{table}[ht]
    \renewcommand\arraystretch{1.7}
    \begin{flushleft}   
    \setlength{\abovecaptionskip}{-0.2cm} 
    \caption{Base System} 
    \centering
	\begin{tabular}{|m{1.8cm}|m{6.4cm}|} 
        \hline Model & Layers \\ 
		\hline \multirow{3}*{Encoder} &Conv2d,kernel=(7,7),stride=(1,1)\\ 
        \cline{2-2} &Conv2d,kernel=(3,3),stride=(2,2) x 4 \\
        \cline{2-2} &Conv2d,kernel=(3,3),stride=(1,1) \\  
        \hline \multirow{4}*{Decoder} &Conv2d,kernel=(3,3),stride=(1,1)\\ 
        \cline{2-2} &RRDB+channelNorm x 9\\
        \cline{2-2} &ConvT,kernel=(3,3),stride=(2,2) x 4 \\
        \cline{2-2} &Conv2d,kernel=(7,7),stride=(1,1) \\
        \hline \multirow{4}*{Discriminator} &For latent Y: nearest neighbor upsampling 16x \\
        \cline{2-2} &concat[upsampled latent Y, input image X or X']\\
        \cline{2-2} &Conv2d,kernel=(3,3),stride=(2,2) x 4 \\
        \cline{2-2} &Conv2d,kernel=(1,1),stride=(1,1) \\
        \hline
	\end{tabular}
\end{flushleft} 
\end{table}
\begin{table}[ht]
    \renewcommand\arraystretch{1.7}
    \begin{flushleft}   
    \setlength{\abovecaptionskip}{-0.2cm}     
    \caption{Enhancement CNN Model} 
	\centering
	\begin{tabular}{|m{1.8cm}|m{6.4cm}|} 
        \hline Model & Layers \\ 
		\hline \multirow{3}*{Enhancement} &Conv2d,kernel=(7,7),stride=(1,1)\\ 
        \cline{2-2} &Conv2d,kernel=(3,3),stride=(1,1) x 8 \\
        \cline{2-2} &Conv2d,kernel=(7,7),stride=(1,1) \\
        \hline
	\end{tabular}
\end{flushleft} 
\end{table}
According to STSCI's system architecture and training process, STSCI is not only suitable for inspecting key components in substations, but also, thanks to its flexible system capabilities, it can be used for monitoring and inspection tasks that focus on specific semantic information in fixed channel conditions. By changing the training data, STSCI can easily change the semantic features it focuses on. For example, by using images of safety facilities such as fire equipment, emergency exits, etc., to fine-tune the system, it can be trained to focus on whether the safety devices are functioning properly. Furthermore, to a certain extent, the system can pay attention to multiple specific semantic elements, such as simultaneously detecting the key components of the substation and ensuring the normal operation of safety equipment. Moreover, STSCI is not limited to the substation scenario. By collecting real channel information from different scenarios to modify the parameters of the simulated-channel model, the system can be adapted to different plant area scenarios. \par 

\section{Experimental results}
This section is mainly introduced the relevant testing settings, including the dataset for STSCI's train and test, the introduction of baseline as well as evalation metrics and the performance for the STSCI in different metrics.\par
Discription and figures are given to show how the STSCI surpass the traditional image communication system or other semantic system under some specific situations.\par
~\\
\emph{A. Dataset for train and test}\par
The training dataset is formed of 10000 images sampled from the COCO2014 dataset while 200 images of substation are used to fine-tune the base system to improve the STSCI's performance in the intelligent substation patrol inspection task.\par
During the testing process, the images from COCO2014 testset which are not involved in training process are sampled to measure the metrics of the communication systems.\par
~\\
\emph{B. Baseline and Evaluation metrics}\par
JPEG and JPEG2000, are used as baselines for image compression.  JPEG is a widely used image compression standard that is commonly employed in image transmission and storage. On the other hand, JPEG2000 is an advanced image compression standard known for its higher compression efficiency and improved image quality.\par
The LSCI proposed in\cite{LSCI} is also used as benchmarks in the comparison, which is a novel generalized semantic communication system. Comparing STSCI and LSCI can demonstrate the superiority of STSCI in handling specific tasks and showcase the effectiveness of the structural optimizations made by STSCI for specific tasks, such as semantic enhancement systems.\par
Meanwhile, the Low-density Parity-check(LDPC) channel coding is used to make comparison with JSCC methods under simulated-channel conditions of the wireless transmission channels, as LDPC codes are powerful error-correcting codes known for their excellent error correction performance and are widely used in digital communication systems. \par
Structural Similarity Index(SSIM) as well as Peak Signal-to-Noise Ratio(PSNR) is chosen as evaluation metrics to measure both the quality of images at the receiving end and the similarity between the transmitted and original ones. PSNR can directly measure the performance of images in terms of signal-to-noise ratio, making it a useful metric in applications that require high fidelity. SSIM considers factors such as image structure and contrast more comprehensively, making it a more suitable metric for applications that require attention to image details. Both metrics can help comprehensively describe the performance of the communication system. \par
\begin{figure*}[ht]
    \centering
    \includegraphics[scale=1]{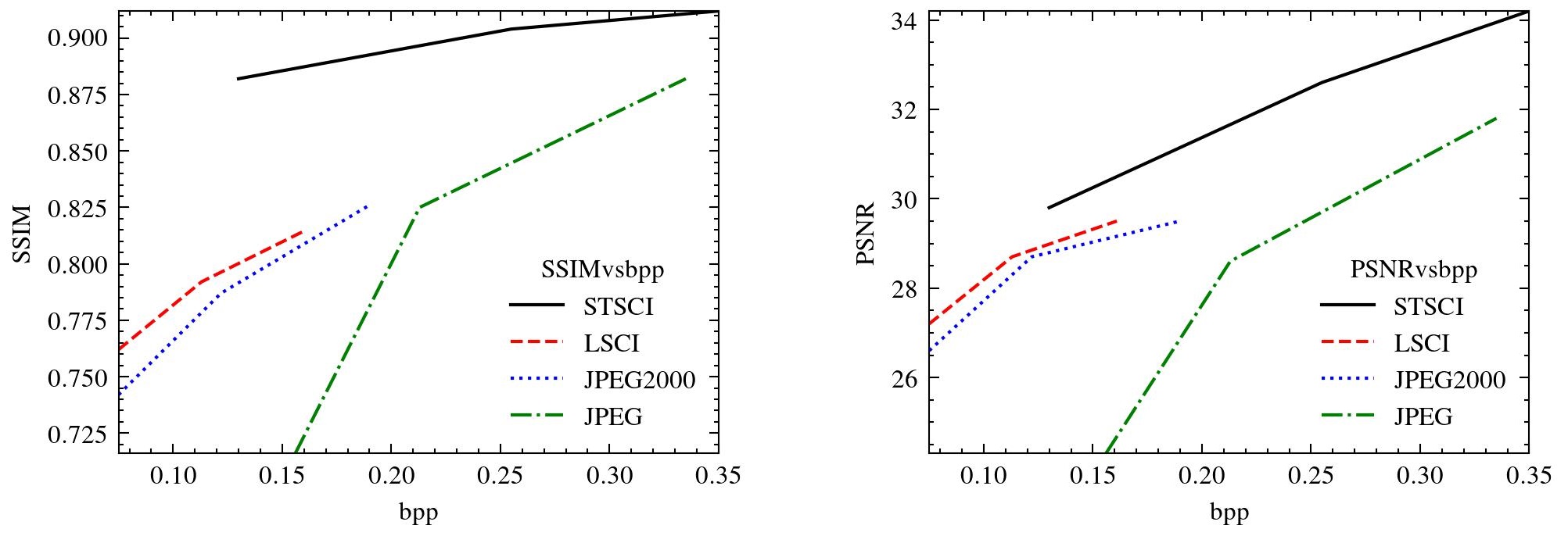}
    \caption{The performance of the reconstructed image of JPEG, JPEG2000, LSCI and STSCI with different bpp. The metrics SSIM and PSNR(dB) measure the distortion performance.}
\end{figure*} 
\begin{figure*}[ht]
    \centering
    \includegraphics[scale=0.30]{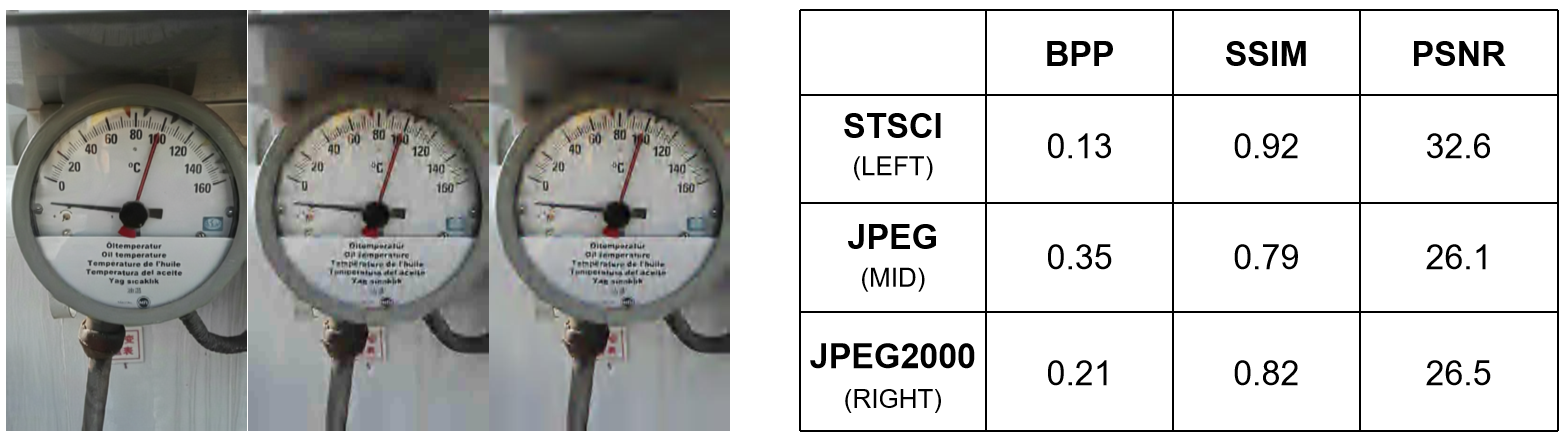}
    \caption{Visual example of images produced by LSCI along with the corresponding results for JPEG and JPEG2000. STSCI has higher metric value in a lower bpp than JPEG and JPEG2000. The image compressed by STSCI is more realistic and the details are clearer.}
\end{figure*}  
\begin{figure}[h]
    \centering
    \includegraphics[scale=0.25]{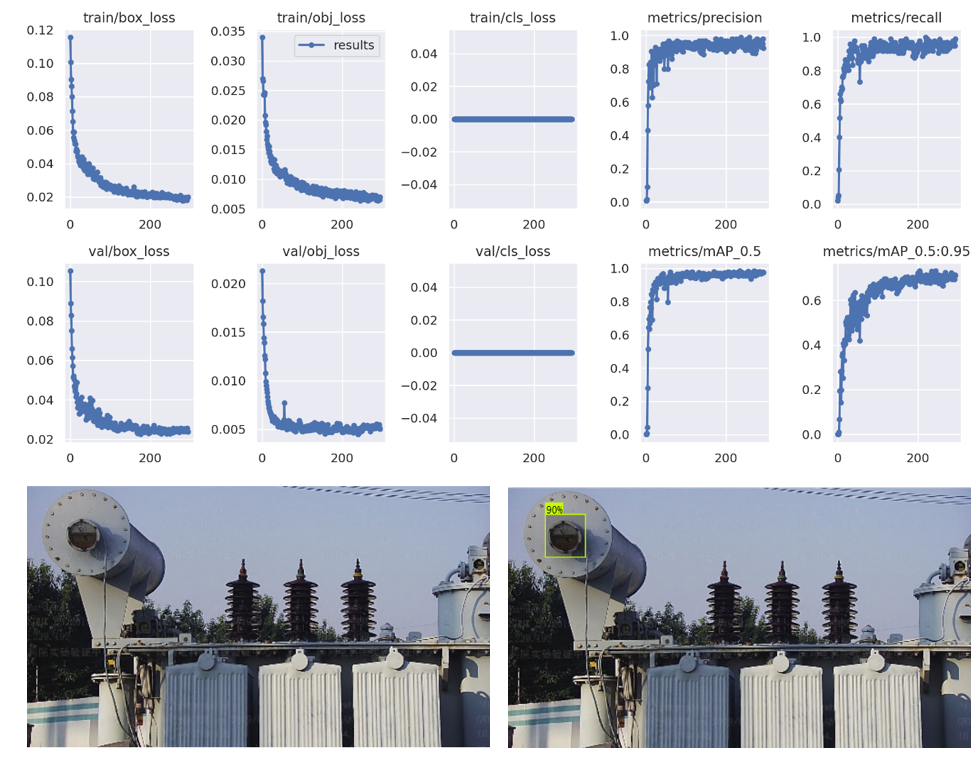}
    \caption{Training details and visual example of the YOLONet. The YoloNet is fine-tuned using images that include instrument panels. The key semantic content area identified by the YoloNet in the test image is highlighted with yellow bounding boxes.}
\end{figure} 
\begin{figure}[h]
    \centering
    \includegraphics[scale=0.2]{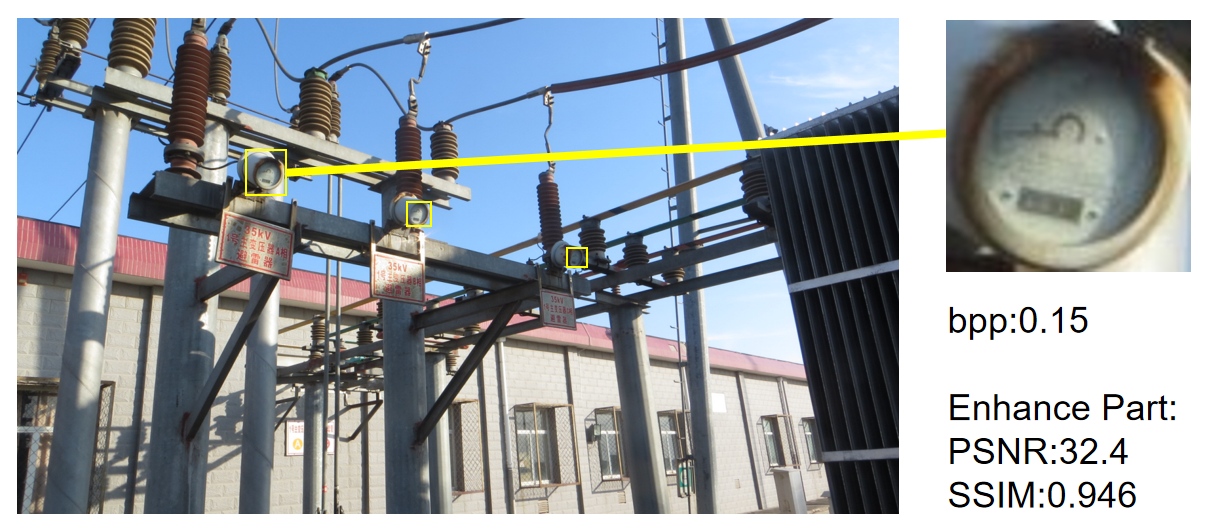}
    \caption{Visual example of the enhancement CNN model. After passing through the enhancement CNN model, the key semantic content areas of the image exhibit improved image metrics. Moreover, visually, the directions of the pointer on the panel is clearly visible.}
\end{figure}  
\begin{figure*}[hb]
    \centering
    \includegraphics[scale=1]{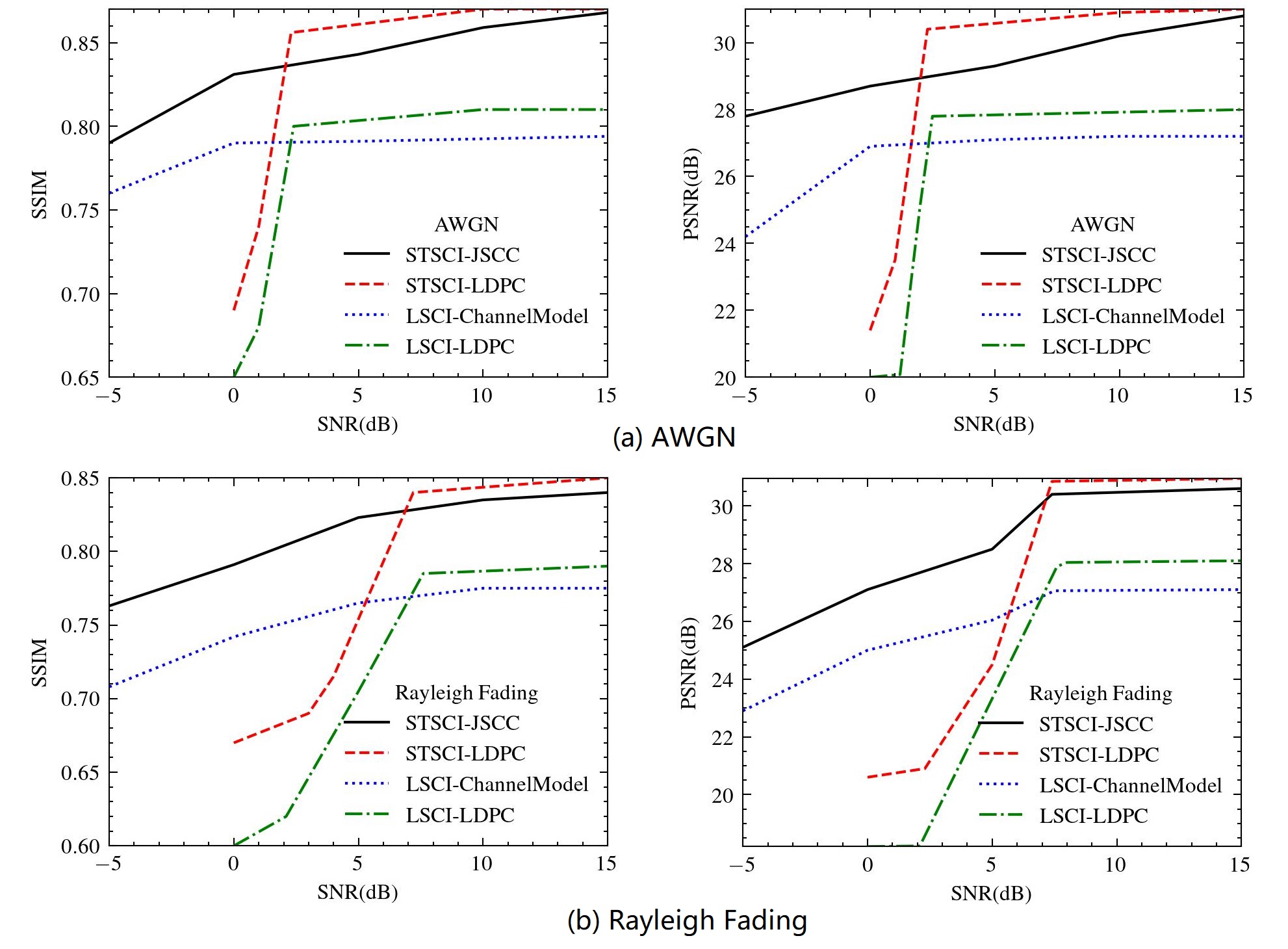}
    \caption{Comparison between STSCI and LSCI with JSCC or channel slice models and traditional channel coding LDPC with SSIM and PSNR metrics in AWGN channel and Rayleigh fading channel. STSCI with JSCC achieves high image metrics and exhibits minimal degradation in performance under low signal-to-noise ratio conditions, maintaining a high level of image transmission quality.}
\end{figure*}  
~\\
\emph{C. Analysis for results in image compression}\par
We visualize the outcome of the comparison between JPEG, JPEG2000, LSCI and STSCI in image compression task in Fig. 4. The x coordinate represents the average bits per pixel (bpp) on the images, while the y coordinate individually show the value of metrics of SSIM and PSNR.\par
From the Fig. 4, it's obvious that STSCI is always preferred to other image compression methods at
equal bitrates. In the bitrate around 0.15, the STSCI is 0.75 higher than the LSCI and JPEG2000 in value of SSIM and 0.75 is a enormous number which means the reconstructed image gained by STSCI is more resemble to the original ones.\par 
And that is extatly the truth, visual examples presented in Fig. 5 shows how clear the imge compressed by the STSCI. Even using only half bpp of JPEG2000 and one of three bpp of JPEG, image handled by STSCI is 0.1 higher in SSIM and around 8dB higher in PSNR metrics. It's esay for us to see noises and distortions in images compressed by JPEG and JPEG2000, compared to which, the STSCI's job is better. Such results in compressing and transmitting the image shows that STSCI can be equal to the specific patrol task with higher quality and less bpp.\par   
Considering that the base system is fine-tuned with some substation and industral images, and that's why in this visual sample, the STSCI's SSIM and PSNR metrics are higher than the average values in 0.13bpp. Indeed, in the substation patrol task, the images of substation can be collected continuously to fine-tune or even retrain the networks of Base system, which can leads to better performance in the specific task.

~\\
\emph{D.Analysis for semantic enhancement system}\par
For example, taking the panel as the key semantic information, a YOLONet is trained with 200 images of panels. Both the details and the example of trained YOLONet is shown in Fig. 6. \par
With pre-trained checkpoints involved, after 200 images' training, the YOLONet is precious enough for the daily patorl task with making errors or missing in low frequency. \par
Meanwhile Fig. 7 shows the effect of the enhancement CNN model. The enhanced area in Fig. 7 has the high SSIM at 0.946 and PSNR at 34.4dB. Through the enhancement CNN model, we can still see the direction of the hand on the panel, which is of great meaningful information for the patrol task. \par
To accurately test the effectiveness of the semantic enhancement system, we conducted experiments using images containing substation units, and limited the bpp of each system to around 0.15. The results were presented in Table III, which shows the average SSIM and PSNR indicators of the output images of each system. The data in Table III indicates that the semantic enhancement system is able to improve the image restoration quality and fidelity of the enhanced area compared to the non-enhanced area, and significantly outperformed traditional JPEG and JPEG2000 image compression methods. These findings demonstrate the protective effect of the semantic enhancement system on key semantic information. By using additional information to perform semantic enhancement operations, the system was able to preserve important semantic information while maintaining a low compression ratio.\par
\begin{table}[h]
    \renewcommand\arraystretch{1.7}
    \begin{flushleft}        
	\centering     
    \caption{Comparison of image metrics between systems} 
	\begin{tabular}{|m{4cm}|m{1.6cm}|m{1.6cm}|} 
        \hline Model & SSIM & PSNR(dB) \\ 
		\hline STSCI(Enhanced Area) & 0.942 & 32.1 \\ 
        \hline STSCI(Non-enhanced Area) & 0.887 & 30.6 \\
        \hline JPEG & 0.714 & 23.9 \\
        \hline JPEG2000 & 0.806 & 29.4 \\ 
        \hline
	\end{tabular}
\end{flushleft} 
\end{table}
~\\
\emph{E.Simulated results for channel communication}\par
In the experiments, we compress the image with bpp around 0.1 and choose AWGN channel as well as Rayleigh fading channel to make channel simulation. The AWGN channel model approximates the noise conditions when the signal is transmitted without obstacles, while the Rayleigh channel incorporates scattering and reflection effects, simulating multi-path and signal fading in wireless channels. These two channel models are widely used for performance evaluation in communication systems, so utilizing these channel models allows for reliable results that validate the system's reliability and adaptability in different channel environments. \par
As shown in Fig. 8, under both AWGN channel and rayleigh fading channel conditions, when the SNR is larger than 5dB, the value of SSIM and PSNR gained by STSCI+LDPC is a litter higher than STSCI+JSCC, but when the channel conditions gets bad and the SNR is close or even lower than 0db, the quality of image transmitted through JSCC metheds doesn't decrease very fast and becomes higher than that of LDPC methods, which is desirable for the specific task.  
One of the most importance mission for STSCI in this task is to ensure the quality of image sent back by robots when patrolling some marginal areas with low signal-to-noise ratio channel conditions. And unlike LSCI whose encder and decoder is not optimized when involving the noise by using channel slice models, STSCI's performance in good channel conditions can get closer and closer to the LDPC metheds. \par
~\\
\begin{table}[b]
    \renewcommand\arraystretch{1.7}
    \begin{flushleft}        
	\centering     
    \caption{System Computational Power Parameters} 
	\begin{tabular}{|m{2.4cm}|m{1.6cm}|m{1.6cm}|m{1.6cm}|}
        \hline Model & Params(M) & Flops(G) & Speed(ms) \\  
        \hline
        \multicolumn{4}{|c|}{Nvidia Orin} \\
        \hline YOLOv5s & 7.2 & 16.50 &  22.6 \\
		\hline Encoder & 6.0 & 34.46 & 50.00 \\ 
        \hline
        \multicolumn{4}{|c|}{Nvidia P100} \\
        \hline Decoder & 6.3 & 49.54 &  19.66 \\
        \hline Enhancement CNN Model & 12.5 & 80.37 & 12.48\\ 
        \hline
	\end{tabular}
\end{flushleft} 
\end{table}
\emph{F.System parameter count, computational power requirements, and running speed}\par
According to Table IV, we can obtain the parameter counts, total computational power, and running speed for each model in the system. The experiments were conducted on an Nvidia Orin at the transimtting and an Nvidia P100 GPU at the receiving end, with image size set to 512x512 pixels(enhanced area are 80x80 pixels). \par
Considering the scenario where the transmitter is an edge device with relatively limited computing power, Nvidia Orin is adopted to perform computational testing on the models deployed at the transmitting end. Nvidia Orin is a chip designed specifically for edge computing, with advantages such as small size and low power consumption, and is widely used in the Internet of Things field, matching the requirements of the scenario. \par
In the model design, we considered that the computational power is limited at the transimtting end while relatively stronger at the receiving end. Therefore, in the design, the complexity of the model is reduced at the transmitting end and improved the ability of the model at the receiving end with more convolutional layers and connections. According to Table IV, the total parameters and flops of the two models at the transimtting end are smaller than those of the models at the receiving end. \par
\begin{figure}[h]
    \centering
    \includegraphics[scale=0.42]{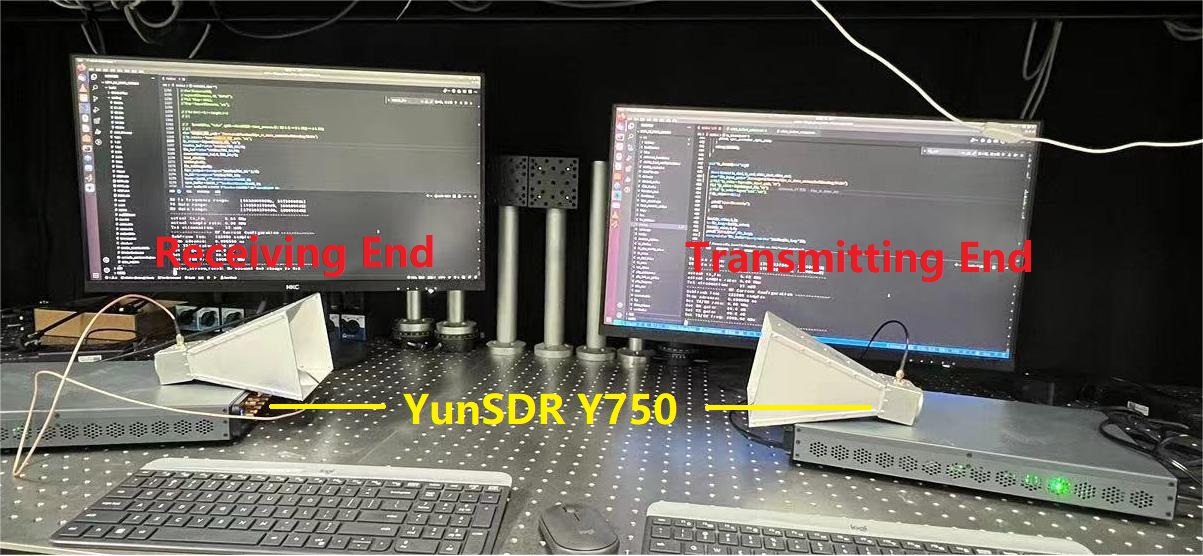}
    \caption{The process of signal transmission and reception in hardware. The transmitting end is located on the right side, while the receiving end is on the left side. The experimental hardware used is the YunSDR Y750.}
\end{figure}  
\emph{G.Hardware Signal Transmission, Reception, and Real Channel Experiments}\par
To further validate the effectiveness and reliability of the system, we deployed the system at both the transmitting and receiving end and conducted real-channel image transmission using hardware. As shown in Fig. 9, YunSDR Y750\footnote{Introduction website : https://www.v3best.com/y750s} devices were used at both the transmitting and receiving ends, with the bitstream employing OFDM modulation and a bpp parameter of 0.1. The measured SNR of the wireless channel was approximately 0dB. After completing the transmission, the average performance metrics of the received images are presented in Table V.\par
\begin{table}[ht]
    \renewcommand\arraystretch{1.7}
    \begin{flushleft}        
	\centering     
    \caption{Hardware Experiment Results Metrics} 
	\begin{tabular}{|m{2.4cm}|m{2.4cm}|m{2.4cm}|}
        \hline  & SSIM & PSNR(dB) \\ 
		\hline Final Image & 0.83 & 27.83 \\ 
        \hline Enhannced Area & 0.88 & 28.77 \\
        \hline
	\end{tabular}
\end{flushleft} 
\end{table}
At a bpp value of 0.1 and an SNR of around 0, the image metrics obtained from the hardware experiment exhibit fluctuations around the results obtained from software simulation. In such low SNR scenarios, our STSCI still performs well both in terms of image metrics and visual effects. The average values are slightly lower but very close to the results from the software simulation. \par
Meanwhile, Fig. 10 provides a visual example of hardware transmission along with its corresponding image metrics. According to Fig. 10, even at SNR around 0dB, the image metrics of the final image are still relatively high, without significant distortion or deformation. In contrast to the blurry and unclear version of the dial without enhancement, the enhanced version maintains clear visibility of the pointers and readings on the dial. \par
These results demonstrate that STSCI is capable of performing well in real hardware deployment and transmitting over real channels. It also confirms the reliability of the software simulation results obtained earlier. \par
\begin{figure*}[ht]
    \centering
    \includegraphics[scale=0.37]{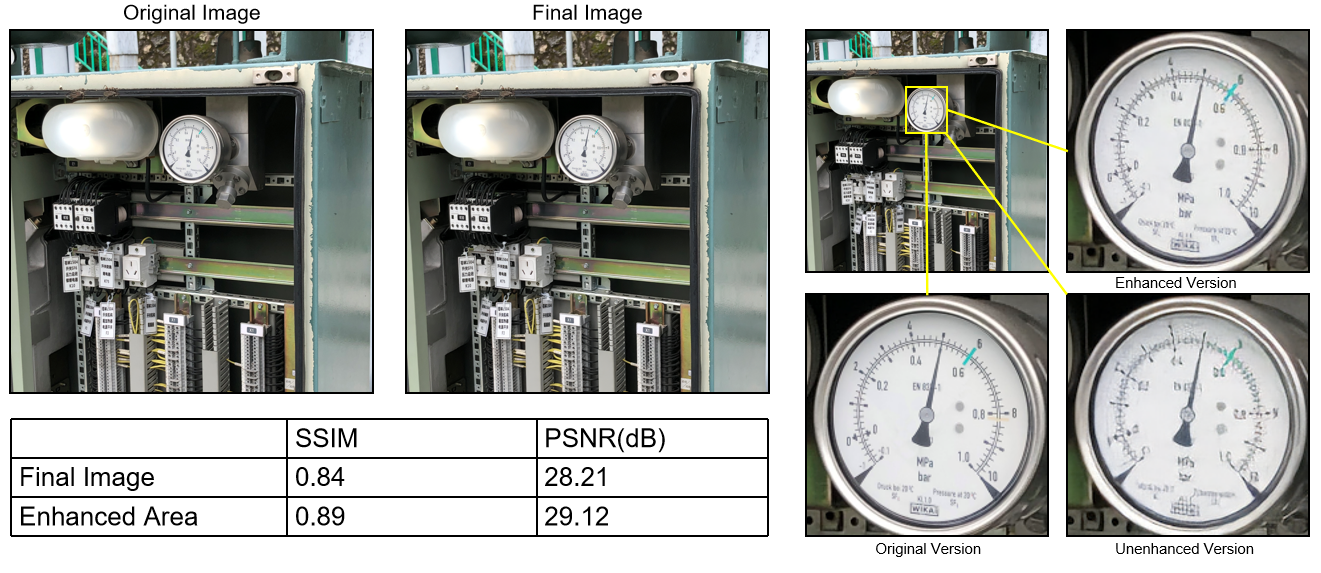}
    \caption{
    Visual examples for hardware signal transmission, reception, and real channel experiments. On the left side, the original image at the transmitting and the final image at the receiving end are displayed. On the right side, the original sub-image of the key semantic region, the transmitted sub-image without enhancement, and the enhanced sub-image are shown. The table contains the image metrics for each section.
    }
\end{figure*}    
\section{Conclusion}
In this paper, a specific task-oriented semantic image communication system STSCI is proposed for intelligent substation patorl inspection. It consists of a base system and a semantic enhancemant system that work together to compress and transmit images with high quality, while preserving the key semantic information. The STSCI utilizes a GAN-based network for image compression, a YOLONet for semantic localization, and an enhancement CNN model to improve the clarity of key semantic areas. JSCC technology is also involved to improve the system's performance under low signal-to-noise ratio channel conditions. \par 
Experimental results show that with all metheds taken, the STSCI has the ability to solve the specific task of intelligent substation patrol inspection, making it a promising technology for future applications in this field. \par

\section{Acknowledgements}
This work is supported in part by the National Key R\&D Program of China under Grant 2022YFB2902102. The work of is supported by The Academician expert Open Fund of Beijing Smart-chip Microelectronics Technology Co., Ltd under project SGITZXDTKJJS2201045, in part by the Fundamental Research Funds for Central Universities(Project Number: 2482021RC01). 

\bibliographystyle{IEEEtran}
\bibliography{new}

\end{document}